\documentclass[nohyperref]{article}

\usepackage{microtype}
\usepackage{graphicx}
\usepackage{subfigure}
\usepackage{booktabs} 

\usepackage{hyperref}

\usepackage[accepted]{icml2022}

\usepackage{amsmath}
\usepackage{amssymb}
\usepackage{mathtools}
\usepackage{amsthm}
\usepackage{bm}
\usepackage{wrapfig}
\usepackage{caption}

\usepackage{multirow}
\usepackage[capitalize,noabbrev]{cleveref}

\theoremstyle{plain}

\theoremstyle{definition}

\theoremstyle{remark}

\usepackage[textsize=tiny]{todonotes}

\icmltitlerunning{Efficient Training of Deep Equilibrium Models}

\begin{document}

\twocolumn[
\icmltitle{Efficient Training of Deep Equilibrium Models}



\icmlsetsymbol{equal}{*}

\begin{icmlauthorlist}
\icmlauthor{Bac Nguyen}{equal,comp}
\icmlauthor{Lukas Mauch}{equal,comp}
\end{icmlauthorlist}
\icmlaffiliation{comp}{Sony Europe B.V., R\&D Center, Stuttgart Laboratory 1, Germany}

\icmlcorrespondingauthor{Bac Nguyen}{bac.nguyencong@sony.com}
\icmlcorrespondingauthor{Lukas Mauch}{lukas.mauch@sony.com}

\icmlkeywords{Deep equilibrium models, implicit layer, efficient training}

\vskip 0.3in
]



\printAffiliationsAndNotice{\icmlEqualContribution} 

\begin{abstract}
Deep equilibrium models (DEQs) have proven to be very powerful for learning data representations. The idea is to replace traditional (explicit) feedforward neural networks with an implicit fixed-point equation, which allows to decouple the forward and backward passes. In particular, training DEQ layers becomes very memory-efficient via the implicit function theorem. However, backpropagation through DEQ layers still requires solving an expensive Jacobian-based equation. In this paper, we introduce a simple but effective strategy to avoid this computational burden. Our method relies on the Jacobian approximation of Broyden’s method after the forward pass to compute the gradients during the backward pass. Experiments show that simply re-using this approximation can significantly speed up the training while not causing any performance degradation.
\end{abstract}

\section{Introduction}
Recently, deep equilibrium models (DEQs) have been successfully applied to a variety of tasks, such as semantic segmentation~\citep{bai2020multiscale}, music source separation~\citep{koyama2022music}, reverse imaging~\citep{gilton2021deep}, image classification, and language modeling~\citep{bai2019deep}. Unlike traditional (explicit) feedforward neural networks, DEQs define their output as the solution to a fixed-point equation. Alternatively, DEQs can be viewed as infinite-depth weight-tied feedforward networks. Training and inference typically rely on fixed-point or root-finding solvers, such as Anderson acceleration~\citep{walker2011anderson} and Broyden's method~\citep{bai2022neural}, for efficient computation during the forward and backward passes.

From a computational point of view, DEQs have some properties that make them worth studying for efficient training and inference. 
1) DEQs decouple the network definition and implementation. While the behavior of the DEQ is defined by the fixed-point equation we can freely choose how to solve it, meaning that we can adapt to the needs of specific platforms. 2) Fixing the tolerance with which we want to find the fixed point makes the computational complexity of the forward pass data-dependent. More specifically, we can often abort the forward iterations after just a few steps in case of simple inputs. In practice, this yields a variable depth network~\cite{bai2019deep}. 3) Compared to explicit models, training DEQs is memory-efficient. According to the implicit function differentiation theorem, for a given input, the gradient of the training loss with respect to the DEQ parameters only depends on the fixed point of the DEQ but not on intermediate forward iterations. Therefore, the memory requirement of DEQ training is constant with the network depth and scales linearly with the layer width~\citep{bai2019deep}. 

Despite these interesting properties, training DEQs is still far from being efficient. The exact gradient computation in the backward pass involves an inverse Jacobian matrix. This could be intractable in high-dimensional features with a cubic time and quadratic memory complexity. To avoid the explicit computation and storage of the Jacobian matrix, previous methods proposed solving a Jacobian-based equation using a black-box root finder. In this way, only vector-Jacobian products are involved~\citep{bai2022neural}, which can be efficiently computed via automatic differentiation. Although the Jacobian matrix is not explicitly constructed, in practice, the backward pass is still computationally expensive since it depends on a number of iterations.

In this paper, we propose a simple approach to avoid this expensive iterative procedure during the backward pass. In particular, we re-use the approximation of the inverse Jacobian matrix given by Broyden’s method~\citep{broyden1965class} after the forward pass to perform the gradient estimation. As a result, the backward pass can be efficiently computed as it only requires a single matrix multiplication. Empirical experiments show that this simple technique can achieve very competitive results while being considerably faster than using a black-box root finder.

\section{Related works} \label{sec:rel}
Despite the success of DEQs in many tasks, reducing their running time is still challenging. This has motivated a very active research area in designing efficient forward and backward algorithms to speed up their training. Many recent works have focused on accelerating the gradient computation during the backward pass~\cite{fung2021jfb,geng2021training}. The idea is to replace the exact gradient computation with an inexpensive gradient approximation. Besides the computational efficiency, approximated gradients can also introduce noises, which have proven to be beneficial for training neural networks~\citep{an1996effects,zhu2018anisotropic,wu2020noisy}.

In particular, \citet{fung2021jfb} proposed Jacobian-Free backpropagation (JFB) to avoid solving the Jacobian-based equation. During the backward pass, it simply replaces the inverse Jacobian-based term with an identity matrix. Under certain conditions, it was shown that JFB gives ascent directions of the loss function in the training process. Similarly, \citet{geng2021training} introduced the Neumann-series-based phantom gradients (NPG) that use a truncated Neumann series to approximate the costly inverse Jacobian-based term. Interestingly, JFB is a zeroth-order approximation to the Neumann series, which can be seen as a special case of NPG. More details will be discussed in Subsection~\ref{sec:sec:efficient}.

\section{Proposed method} \label{sec:gdeq}
In this section, we first review DEQs and then discuss how to apply the implicit function theorem for the gradient computation. Because implicit differentiation requires an iterative procedure to solve a system of linear equations, it makes the gradient computation very slow. To reduce this computational burden, we propose GDEQ, a simple method to compute the backward pass which only involves a simple matrix computation. As a result, it yields a faster training scheme and a simpler implementation.

\subsection{Deep equilibrium models}
Let $f_{\bm{\theta}}(\mathbf{z}, \mathbf{x})$ parameterized by $\bm{\theta} \in \mathbb{R}^{d_{\bm{\theta}}}$
be a DEQ layer with input $\mathbf{x} \in \mathbb{R}^{d_\mathbf{x}}$ and state $\mathbf{z} \in \mathbb{R}^{d_\mathbf{z}}$.
Furthermore, let $\mathbf{z}^*$ be the equilibrium point of $f_{\bm{\theta}}$ that satisfies
\begin{align}
    g_{\bm{\theta}} (\mathbf{z^{*}}, \mathbf{x}) &:= f_{\bm{\theta}}(\mathbf{z^{*}}, \mathbf{x}) - \mathbf{z^{*}} = \mathbf{0}\,. \label{eq:forward}
\end{align}
Once the equilibrium is found, it is passed to another post-processing layer to compute the prediction, $\mathbf{\hat{y}} = \mathcal{F}(\mathbf{z^{*}}) \in \mathbb{R}^q$. Let $\mathcal{L}: \mathbb{R}^q \times \mathbb{R}^q \to \mathbb{R}$ be a differentiable
function, the network is trained based on the following expected loss
\begin{align*}
    \ell = \mathbb{E}_{(\mathbf{x},y)} [\mathcal{L}(\mathbf{\hat{y}}, \mathbf{y})]\,,
\end{align*}
where $\mathbf{y}\in \mathbb{R}^q$ indicates the ground-truth corresponding to the training example $\mathbf{x}$.

\textbf{Forward pass.} 
For a given input $\mathbf{x}$, the goal is to find the equilibrium point $\mathbf{z^{*}}$ by solving Eq.~(\ref{eq:forward}). Broyden’s method~\citep{broyden1965class} is used to iteratively update the estimation of the equilibrium. More specifically, let $\mathbf{z}_0$ be an initial random state of the equilibrium layer. At the $t$-th iteration, we apply the following update rule
\begin{align*}
    \mathbf{z}_{t + 1} = \mathbf{z}_{t} - \mathbf{B}^{-1}_{t} g_{\bm{\theta}} (\mathbf{z}_{t}, \mathbf{x})\,,
\end{align*}
where $\mathbf{B}^{-1}_{t}$ denotes an approximation of the inverse Jacobian matrix $\nabla^{-1}_{\mathbf{z}}g_{\bm{\theta}} (\mathbf{z}_{t}, \mathbf{x})$ at $\mathbf{z}_t$. At the beginning, $\mathbf{B}^{-1}_{0}$ is set to be $-\mathbf{I}$. To keep this matrix close to the Jacobian, we update it after each iteration
using the Sherman-Morrison formula~\citep{sherman1950adjustment}
\begin{align*}
    \mathbf{B}^{-1}_{t + 1} = \mathbf{B}^{-1}_{t} - \frac{\mathbf{B}^{-1}_{t}g_{\bm{\theta}} (\mathbf{z}_{t + 1}, \mathbf{x})\Delta \mathbf{z}_{t + 1}^\top\mathbf{B}^{-1}_{t}}{\Delta \mathbf{z}_{t+1}^\top\Big(\Delta \mathbf{z}_{t+1} + \mathbf{B}^{-1}_{t} g_{\bm{\theta}} (\mathbf{z}_{t + 1})\Big)}\,,
\end{align*}
where $\Delta \mathbf{z}_{t+1} = \mathbf{z}_{t+1} - \mathbf{z}_{t}$. The algorithm stops when a maximum number of iterations is reached or a solution is found.

\textbf{Backward pass.} To train the DEQ, we need to calculate gradients with respect to $\bm{\theta}$ and $\mathbf{x}$. Using the implicit function theorem~\citep{krantz2002implicit}, we can compute the gradients with respect to the model parameters $\bm{\theta}$ as
\begin{align}
\frac{\partial \ell}{\partial \bm{\theta}} = \frac{\partial f}{ \partial \bm{\theta}} \mathbf{A} \frac{\partial \ell}{\partial \mathbf{z^{*}}}\, \label{eq:backward}   
\end{align}
where 
\begin{align*}
    \mathbf{A} = \left(\mathbf{I} - \frac{\partial f}{ \partial \mathbf{z^{*}}}\right)^{-1}\,.
\end{align*}
A similar formulation can be derived to compute the gradients of $\ell$ with respect to the input representation $\mathbf{x}$. The main computational burden in Eq.~(\ref{eq:backward}) is dominated by 
the calculation of $\mathbf{A}$. It quickly becomes intractable when $d_{\mathbf{z}}$ is large. This is due to the cubic time complexity of the inverse term as well as the quadratic memory complexity of the Jacobian matrix. 

\subsection{Efficient backward pass} \label{sec:sec:efficient}
The goal is to compute Eq.~(\ref{eq:backward}) in an efficient manner.
Several simplifications have been proposed. 
More specifically, JFB~\citep{fung2021jfb} simply ignores the Jacobian inverse and uses $\mathbf{A} = \mathbf{I}$. This dramatically reduces the computational complexity. NPG~\citep{geng2021training} uses a truncated Neumann series to approximate the inverse, yielding
\begin{align*}
    \bf{A} &= \lambda \sum_{i=0}^{k-1} \mathbf{M}^i \,,\\
    \bf{M} &= \lambda \frac{\partial f}{ \partial \mathbf{z^{*}}} + (1-\lambda) \bf{I}\,,
    \label{eq;npg}
\end{align*}
where $k \in \mathbb{N}$ is the sequence length and $\lambda \in [0,1]$ is a dampening factor. Instead of calculating  $\left(\mathbf{I} - \frac{\partial f}{ \partial \mathbf{z^{*}}}\right)^{-1}$, we only have to compute $k-1$ vetor-Jacobian products. Typically, we choose $k=5$ and $\lambda = 0.5$ for optimal performance~\citep{geng2021training}.

To further reduce the computation, we propose to compute Eq.~(\ref{eq:backward}) by re-using the approximation of the inverse Jacobian term $\mathbf{B}_T^{-1}\approx - \left(\mathbf{I} - \frac{\partial f}{ \partial \mathbf{z^{*}}}\right)^{-1}$ given by Broyden's method at the last iteration after the forward pass. This can avoid the intractable computation of the exact $\mathbf{A}$ in Eq.~(\ref{eq:backward}). More specifically, the gradient computation is formulated as
\begin{align*}
    \frac{\partial \ell}{\partial \bm{\theta}} \approx -\frac{\partial f}{ \partial \bm{\theta}} \mathbf{B}_T^{-1} \frac{\partial \ell}{\partial \mathbf{z^{*}}}\,.
\end{align*}
The main advantage is that the gradient computation in the backward pass only involves a simple matrix multiplication. This is much simpler than solving a system of linear equations as proposed by~\citet{bai2019deep}. Besides the computational simplicity, it can also avoid additional errors that would be introduced by running an additional root-finding algorithm, as proposed in the original DEQ paper. Note that the matrix $\mathbf{B}^{-1}_T$ is written as a sum of $\mathbf{B}^{-1}_0$ and low-rank updates $\mathbf{B}^{-1}_T = \mathbf{B}^{-1}_0 + \sum_{i=1}^T \mathbf{u}_i\mathbf{v}_i^\top$. When the maximum memory storage $m$ is reached, we keep the last $m$ low-rank updates and discard the oldest updates~\citep{mdeq}. We refer to the proposed method as GDEQ.

In addition, Table~\ref{tab:memory} reports the time and memory complexities of different methods required for the backward pass. Here $K$ and $k$ denote the number of iterations used in Broyden's method and the number of unrolling steps used in NPG, respectively. As shown in the table, JFB and GDEQ are the most efficient since they both enjoy constant training time and memory complexities.

\begin{table}[!t]
\caption{Comparison of time and memory complexities among the competing methods.}
\label{tab:memory}
\begin{center}
\begin{tabular}{lcc}
\toprule
Method & Time  & Memory  \\
\midrule
Implicit  & $\mathcal{O}(K)$ & $\mathcal{O}(1)$  \\
JFB & $\mathcal{O}(1)$ & $\mathcal{O}(1)$  \\
NPG & $\mathcal{O}(k)$ & $\mathcal{O}(1)$ \\
\midrule
GDEQ & $\mathcal{O}(1)$ & $\mathcal{O}(1)$\\
\bottomrule
\end{tabular}
\end{center}

\end{table}

\section{Experiments} \label{sec:experiment}
In this section, we conduct experiments on the CIFAR-10 dataset for image classification to show the effectiveness of our method.
Multiscale deep equilibrium models (MDEQ)~\citep{mdeq} is used as the backbone model\footnote{Code available at \url{https://github.com/locuslab/deq}}. All experiments run on 4 NVIDIA RTX A6000 GPUs and use the same configuration.

In Table~\ref{tab:result}, we report the comparison between \textbf{GDEQ} and other state-of-the-art methods, namely with 1) implicit differentiation that solves the system of linear equations during the backward pass (\textbf{Implicit})~\citep{bai2019deep}, 2) Jacobian-free backpropagation (\textbf{JFB})~\citep{fung2021jfb}, and 3) the Neumann-series-based phantom gradient method (\textbf{NPG})~\citep{geng2021training}. Following~\citet{bai2020multiscale}, we use Broyden's method with 18 iterations in the forward and 20 iterations in the backward pass.

As seen from Table~\ref{tab:result}, GDEQ can significantly speed up the training time. It is 2.16x faster than the implicit differentiation.  In an early stage of training, it is common to explicitly unroll the DEQ layers as a weight-tied feedforward neural network. The pre-trained weights are then used to initialize the parameters of MDEQ. In our experiment, we report the classification accuracies under both settings, with and without the pre-training step. The weight initialization during the pre-training stage helps to stabilize the training of MDEQ. It consistently improves the classification accuracies for most of the competing methods. We plan to investigate why this happens in the future.

\begin{table}[!t]
\caption{Performances of the competing methods on the CIFAR-10 dataset. We report the training speed and the classification accuracy.}
\label{tab:result}
\begin{center}
\begin{tabular}{lccc}
\toprule
\multirow{2}{*}{Method} & \multirow{2}{*}{Speedup}& \multicolumn{2}{c}{Accuracy} \\
\cmidrule{3-4}
&  &   without & with\\
\midrule
Implicit   & 1.00x & 91.64 & 92.29 \\
JFB  & 1.97x & 91.56 & 91.99 \\
\text{NPG} &  1.71x & \textbf{93.42} & 92.46 \\
\midrule
GDEQ   & \textbf{2.16x}& 92.01 & \textbf{93.08}  \\
\bottomrule
\end{tabular}
\end{center}

\end{table}

\begin{figure}[!t]
    \centering
	{\includegraphics[trim=50 290 50 290,clip=true, width=0.9\linewidth]{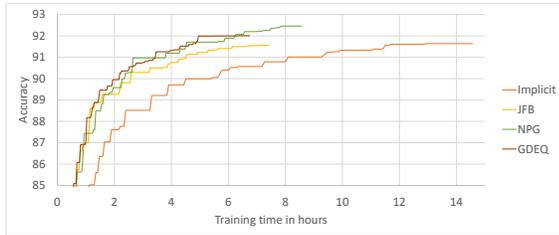}}
    \caption{Convergence of the MDEQ on the test set for different backward algorithms, using no pre-training.}
    \label{figure:convergence}
\end{figure}

Figure~\ref{figure:convergence} compares the convergence speed of the MDEQ with different backward algorithms. As before, we consider implicit differentiation, JFB, NPG, and our GDEQ. We show the accuracy over training time. Obviously, GDEQ converges fastest, performing $220$ training epochs in just $6.7$ hours. This is about 2.16x faster than the vanilla DEQ training that uses implicit differentiation, which takes $14.6$ hours for $220$ epochs. Note, that this is even faster than with JFB, although the backward pass is slightly more expensive. We explain this with the slower convergence caused by a less accurate gradient estimation and the requirement of more forward iterations associated with it. However, given more time GDEQ is outperformed by NPG. 

\begin{figure}[!t]
    \centering
	\subfigure[Epoch 1]
	{\includegraphics[width=0.48\linewidth]{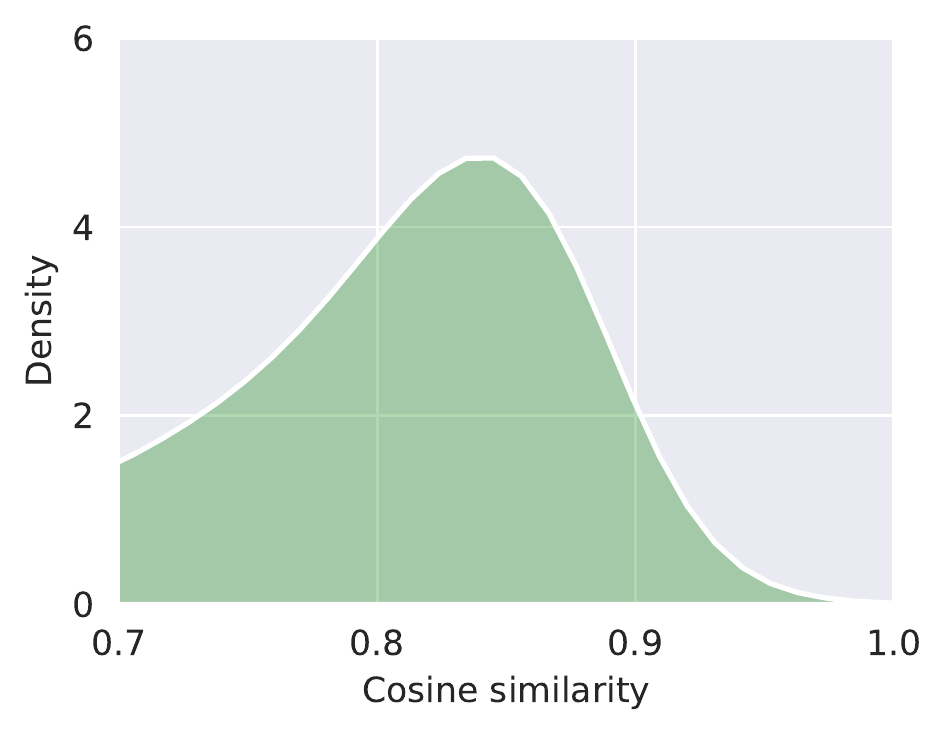}}
	\subfigure[Epoch 10]
	{\includegraphics[width=0.48\linewidth]{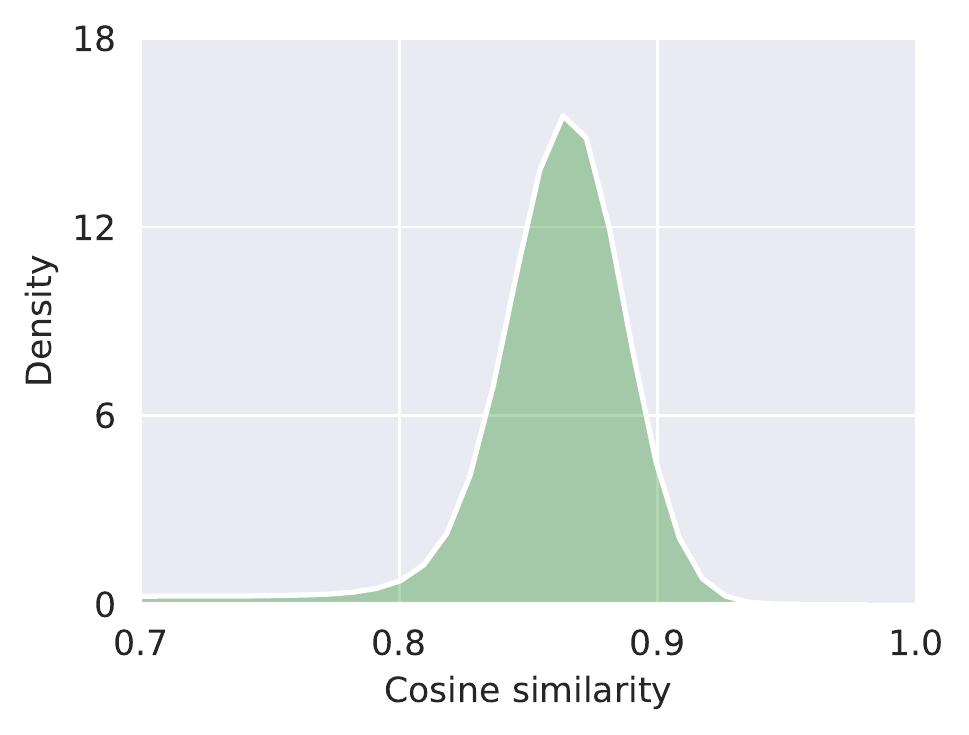}}
	\subfigure[Epoch 20]
	{\includegraphics[width=0.48\linewidth]{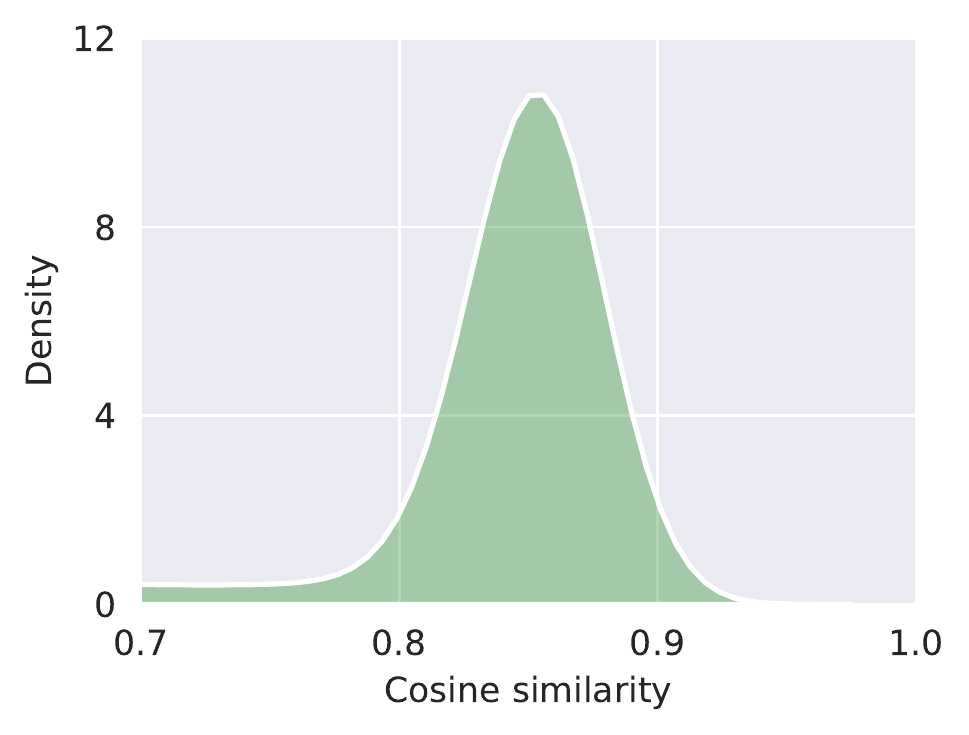}}
	\subfigure[Epoch 50]
	{\includegraphics[width=0.48\linewidth]{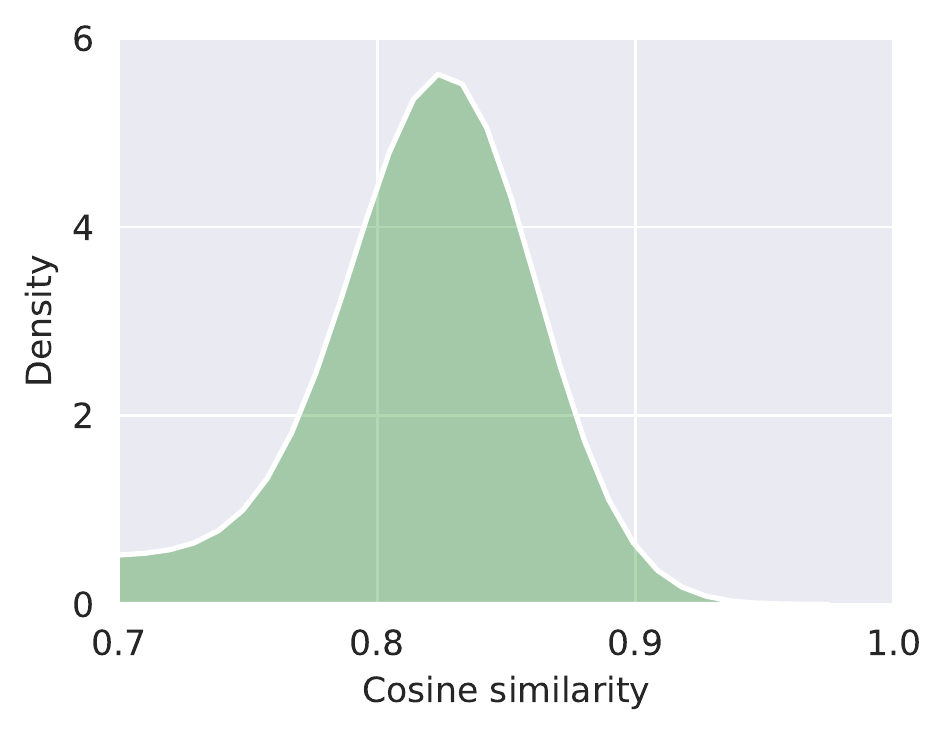}}
    \caption{Cosine similarity between the gradients computed by our method and implicit differentiation at different training epoch.}
    \label{figure:simialrity}
\end{figure}

Furthermore, we perform an ablation study to see how similar the gradients approximated by \textbf{GDEQ} are to those computed by \textbf{Implicit}. Figure~\ref{figure:simialrity} illustrates the cosine similarity of gradients computed by the two methods. It shows that the approximation of the inverse Jacobian given by Broyden's method is quite close to that of Implicit. This implies that GDEQ can still provide the ascent direction of the loss function. 

\section{Conclusions}
In this paper, we have demonstrated that DEQs can be efficiently trained by using an approximation of the inverse Jacobian matrix given by Broyden's method during the forward pass. Several studies on image classification tasks have been conducted to verify our claim. In particular, the proposed method GDEQ achieves 2.16x faster in terms of training time, while being competitive in terms of classification performances. Clearly, it shows great advantages for training DEQs on large-scale datasets. 

\textbf{Limitations.} Despite the promising results, our work has some limitations. First, it lacks studies when Broyden's method does not converge during the forward pass under a limited number of iterations. Since the proposed method heavily depends on the approximation given by Broyden's method, a very biased approximation may lead to unexpected results. Second, it lacks a theoretical convergence guarantee of stochastic gradient descent using gradients provided by our method. Finally, it is also interesting to extend our studies of GDEQ to other task domains rather than image classification. Our future work will aim to address these issues.

\bibliography{example_paper}
\bibliographystyle{icml2022}

\end{document}